\documentclass{article}

\PassOptionsToPackage{numbers, compress}{natbib}
\usepackage[preprint]{nips_2018}

\usepackage[utf8]{inputenc} % allow utf-8 input
\usepackage[T1]{fontenc}    % use 8-bit T1 fonts
\usepackage{hyperref}       % hyperlinks
\usepackage{url}            % simple URL typesetting
\usepackage{booktabs}       % professional-quality tables
\usepackage{amsfonts}       % blackboard math symbols
\usepackage{nicefrac}       % compact symbols for 1/2, etc.
\usepackage{microtype}      % microtypography
\usepackage{graphicx}
\usepackage{siunitx}

\title{Image-based Natural Language Understanding\\Using 2D Convolutional Neural Networks}

\author{Erin\c c Merdivan\thanks{The authors contributed equally to this work.
Corresponding author \texttt{erinc.merdivan@ait.ac.at}} \thanks{Department of Health \& Environment - Austrian Institute of Technology,GmbH}
\thanks{CentraleSupélec, Université de Lorraine, CNRS, LORIA, F-57000 Metz, France}
\and 
\textbf{Anastasios Vafeiadis}\footnotemark[1] \thanks{Information Technologies Institute (ITI) - Center for Research \& Technology Hellas (CERTH)} \and 
\textbf{Dimitrios Kalatzis}\footnotemark[1] \footnotemark[4] \and 
\textbf{Sten Hanke}\footnotemark[2] \and 
\textbf{Johannes Kropf}\footnotemark[2] \and
\textbf{Konstantinos Votis}\footnotemark[4] \and
\textbf{Dimitrios Giakoumis}\footnotemark[4] \and 
\textbf{Dimitrios Tzovaras}\footnotemark[4] \and
\textbf{Liming Chen}\thanks{De Montfort University (DMU)} \and
\textbf{Raouf Hamzaoui}\footnotemark[5] \and
\textbf{Matthieu Geist}\thanks{Université de Lorraine, CNRS, LIEC, F-57000 Metz, France (now at Google Brain)}}

\begin{document}

\maketitle

\begin{abstract}
We propose a new approach to natural language understanding in which we consider the input text as an image and apply 2D Convolutional Neural Networks to learn the local and global semantics of the sentences from the variations of the visual patterns of words. Our approach demonstrates that it is possible to get semantically meaningful features from images with text without using optical character recognition and sequential processing pipelines, techniques that traditional Natural Language Understanding algorithms require. To validate our approach, we present results for two applications: text classification and dialog modeling. Using a 2D Convolutional Neural Network, we were able to outperform the state-of-art accuracy results of non-Latin alphabet-based text classification and achieved promising results for eight text classification datasets. Furthermore, our approach outperformed the memory networks when using out of vocabulary entities from task 4 of the bAbI dialog dataset.
\end{abstract}

\section{Introduction}

Recent advances in natural language processing make heavy use of neural network models. Solutions for tasks such as semantic tagging \cite{collobert2008unified}, text classification \cite{sebastiani2002machine} and sentiment analysis \cite{collobert2011natural} rely on either Recurrent Neural Network (RNN) or Convolutional Neural Network (CNN) variants. In the latter case, the vast majority of the proposed models are based on character-level CNNs applied on one-hot vectors of text or 1D CNNs \cite{huang2016character}. Although the results are promising, having either surpassed or equaled the previous state of the art, there are a few issues regarding the proposed models, which are all related to the fundamental inductive bias underlying these models' architectural design. Whether working at the word- or character-level, language processing with most neural network models almost always translates to sequential processing of a string of abstract discrete symbols. 

CNNs based on 1D or character convolutions constitute the vast majority of CNN models used in language processing. These networks are fast if the dictionary size is small. However, for some languages, the one-hot encoding vector dimension for input sequences can be very large (e.g., over 3000 for Chinese characters). Furthermore, and specifically for RNN variants, training for long input sequences is difficult due to the well-known problem of vanishing gradients. While architectures like Long Short-Term Memory (LSTM) \cite{hochreiter1997long} and Gated Recurrent Units (GRU) \cite{cho2014learning} were specifically designed to tackle this problem, stable training on long sequences remains an elusive goal, with recent works devising yet more ways to improve performance in recurrent models \cite{seo2017neural, trinh2018learning,yu2017learning}. Moreover, many state of the art recurrent models rely on the attention mechanism to improve performance \cite{bahdanau2014neural,luong2015effective,vaswani2017attention}, which places an additional computational burden on the overall method. 

To tackle the above problems, we use CNNs to process the entire text at once as an image. In other words, we convert our textual datasets into images of the relevant documents and apply our model on raw pixel values. This allows us to sensibly apply 2D convolutional layers on text, taking advantage of advances in neural network models designed for and targeting computer vision problems. Doing so, allows us to bypass the issues stated earlier relating to the use of 1D character-level CNNs and RNNs, since now the processing of documents relies on parallel extraction of visual features of many lines (depending on filter size) of text. As far as the vanishing/exploding gradient is concerned, for large CNN architectures, we can easily take advantage of recent architectural advances \cite{he2016deep,he2016identity,chollet2017xception}, which specifically aim to improve its effects. In terms of linguistics, our approach is based on the distributional hypothesis \cite{harris1954distributional}, where our model produces compositional hierarchies of document semantics by way of its hierarchical architecture. Beyond providing an alternative computational method to deal with the problems described above, our approach is also motivated by findings in neuroscience, cognitive science and the medical sciences where the link between visual perception and recognition of words and semantic processing of language has long been established \cite{koelsch2004music, simos1997source}. Our approach is robust to textual anomalies, such as spelling mistakes, unconventional use of punctuation (e.g., multiple exclamation marks), etc. which factors in during feature extraction. As a result, not only is the need of laborious text preprocessing removed, but the derived models are able to understand the semantic significance of the occurrence of such phenomena (e.g., multiple exclamation marks to denote emphasis), which proves to be especially helpful in tasks such as text classification and/or sentiment analysis. Moreover, our approach can work with any alphabet (latin and non-latin), text font, misspellings and punctuation. Furthermore, it can be extended to handwriting, background colors and table formatted text naturally. It also removes the need of pre-processing real-world documents (and thus the need for optical character recognition, spell check, stemming, and character encoding correction).

Our approach is based on the hypothesis that more semantic information can be extracted from features derived from the visual processing of text than by processing strings of abstract discrete symbols. We test this hypothesis on Natural Language Processing (NLP) tasks and show that a solid understanding of text semantics leads to better model performance. Our contributions are summarized as follows:

\begin{itemize}
\item a proof of concept that text classification can be achieved over an image of the text;

\item a proof of concept that basic dialogue modeling (restaurant booking), in an information retrieval setting, can be completed using only image processing methods;
\end{itemize}

The remainder of the paper is organized as follows: Section 2 positions our approach compared to related work, Section 3 introduces the proposed method, Section 4 presents the experimental results and Section 5 draws the conclusions.

\section{Related Work}
\label{gen_inst}

The use of convolutional neural networks for natural language processing has attracted increasing attention in recent years. For sentence classification, Kim \cite{kim2014convolutional} used a simple CNN architecture consisting of one convolutional layer with multiple filters of different sizes, followed by max-pooling. The feature maps produced are then fed to a softmax layer for classification. Despite its simplicity, this architecture exhibited good performance. Sentence modeling was further explored by Blunsom et al. \cite{blunsom2014convolutional} who used an extended application, which they call Dynamic 
Convolutional Neural Network (DCNN) to deal with various input lengths and short- and long-term linguistic dependencies. Wang et al. \cite{wang2015semantic} perform clustering in an embedding space to derive semantic features which they then feed to a CNN with a convolutional layer, followed by k-max pooling and a softmax layer for classification. 

Character-level (as opposed to word- or sentence-level) feature extraction was investigated by 
Zhang et al. \cite{zhang2015character} who used a standard deep convolutional architecture for text classification. Dos Santos and Gatti \cite{dos2014deep} carried out sentiment analysis on sentences taken from text corpora, using a CNN architecture which derives input representations that are hierarchically built from the character to the sentence level. Johnson and Zhang \cite{johnson2014effective} used a CNN for text categorization. Their method does not rely on pre-trained word embeddings, but rather computes convolutions directly on high-dimensional text data represented by one-hot vectors. An architectural variation 
was also proposed for adapting a bag-of-words model in the convolutional layers. Johnson and 
Zhang \cite{johnson2015semi} used CNNs for sentiment and topic classification in a semi-supervised framework, where they retained the representations derived by a CNN over text regions, and which they then integrated into the supervised CNN classifier. Ruder et al. \cite{ruder2016character} employed a novel architecture combining character- and word-level channels to determine an unseen text's author among a large number of potential candidates, a task they called large-scale authorship attribution. Bjerva et al. \cite{bjerva2016semantic} introduced a semantic tagging method, which combines (a) stacked neural network models, consisting of a vanilla CNN or a ResNet \cite{he2016identity} in the lower level for character-/word-level feature extraction and a bidirectional Gated Recurrent Unit (GRU) in the higher level, with (b) a residual bypass function which preserves the saliency of lower-level features that could be potentially lost in the processing chain of intermediate layers. 

While all the aforementioned works used CNNs for NLP tasks, all essentially used text data as input, either pre-trained word embeddings or simply one-hot vector representations.

\section{Method}

\begin{figure*}[!ht]
\begin{center}
   \includegraphics[width=0.9\linewidth]{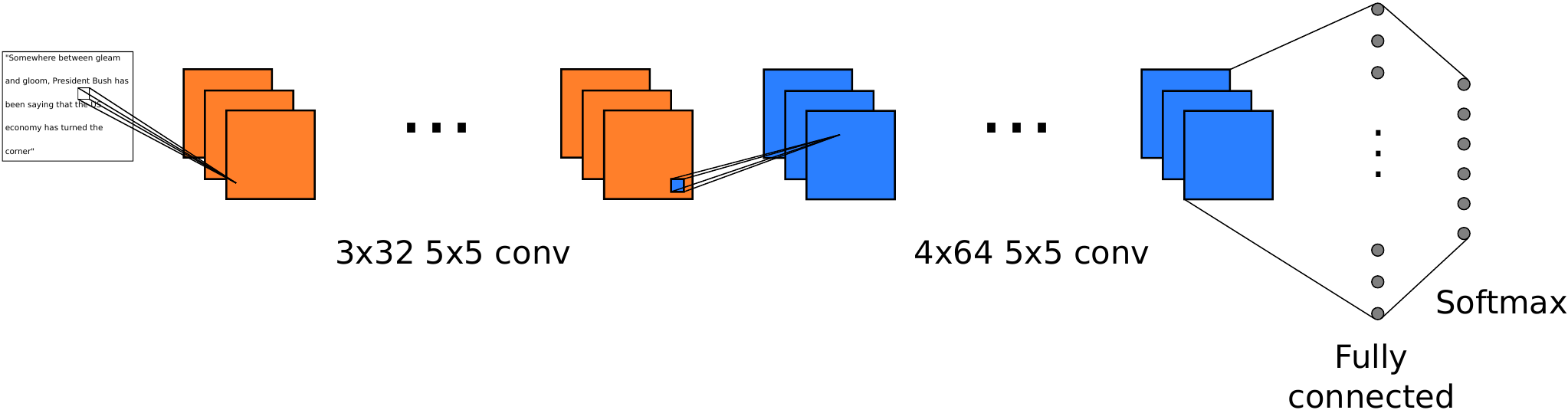}
\end{center}
   \caption{Proposed model: 3 convolutional layers consisting of 32 5x5 filters each, are followed by 4 
   convolutional layers consisting of 64 5x5 filters each. A linear fully connected layer and a classification 
   output layer complete the model.}
\label{fig:architecture}
\end{figure*}

In our approach, we treat text understanding as a problem which concerns the learning of context-dependent semantic conventions of language use in a given corpus of text. We treat this complex problem as an image processing problem, where the model processes an image with the text body (Figure \ref{fig:data_images}), learning both the local (word- and sentence-level) and the global semantics of the corpus. In this way, the domain or context dependent meaning of sentences is implicitly contained in the variations of the visual patterns given by the distribution of words in sentences. As such, the problem reduces to a single, where the model needs to observe as many variations of in-domain text as possible to be able to generalize adequately. This process is similar to the analytical method of learning to read \cite{cebe2011apprendre}, where the global meaning of a body of text is acquired first and learning of the text's meaning moves to hierarchically lower linguistic units. In our case, this translates to understanding the structure and context of the whole corpus first, then the sentences, and finally the words that constitute these sentences.

\subsection{Models} \label{sec:models}
For the tasks of (English and Chinese) text classification we used a vanilla CNN and Xception architecture \cite{chollet2017xception}  to compare 
if better vision deep networks can increase performance. We call this network architecture Xception in paper. 

The vanilla CNN consists of 7 convolutional layers, a fully connected 
layer and an output layer containing as many units as classes (e.g., for a classification problem with 4 classes, the output layer would contain 4 units). All filters in the convolutional layers are 5x5 with stride 2. The first 3 layers use 32 filters, while the rest use 64 filters. The fully connected layer consists of 128 units. All units in all layers use the rectifier function, apart from the output layer, which uses a softmax output. Figure \ref{fig:architecture} shows the architecture of our model.

For the task of dialog modeling we used version 4 of the recently proposed deep Inception network (Inception-V4) \cite{szegedy2017inception}. Our choice was motivated by the fact that the vanilla CNN model was too simple to effectively model the dialog structure, as well as its pragmatics (i.e., the use of language in discourse within the context of a given domain), a problem which Inception-V4 seems to have tackled, at least to a certain extent. We selected the Inception-V4 against the Xception because we wanted to experiment with different advanced architectures for similar tasks.

\subsection{Data Augmentation} \label{sec:augmentation}
Data augmentation has been shown to be essential for training robust models \cite{conneau2017very,copestake1997augmented}. For image recognition, augmentation is applied using simple transformations such as shifting the width and the height of images by a certain percentage, scaling, or randomly extracting sub-windows from a sample of images \cite{de2016data}.

For the task of English and Chinese text classification, we used the {\it ImageDataGenerator} function provided by Keras. The input image was shifted in width and height by 20\%, rotated by $\ang{15}$ and flipped horizontally, using a batch size of 50. For the task of dialogue modeling, we applied the same augmentation techniques and random character flipping. Character flip and in particular changing the rating of a restaurant improved the per-response and per-dialog accuracy, especially for difficult sentences, such as booking a 4 star restaurant.

\section{Results}
To validate our approach, we ran experiments for two separate tasks: text classification 
and dialog modeling, using a single NVIDIA GTX 1080 Ti GPU.

\subsection{Text classification}
In this task we trained our model on an array of datasets which contained text related to news 
(AG's News and Sogou's News), structured ontologies on Wikipedia (DBPedia), reviews (Yelp and Amazon) and question answering (Yahoo! answers). Details about the datasets can be found in \cite{zhang2015character}. For this task, Zhang et al. \cite{zhang2015character} tested CNNs that use 1D convolutions in the task of text classification, which may more broadly include natural language understanding, as well as sentiment analysis. While the model used in \cite{zhang2015character} uses text as input vectors, our proposed method uses image data of text. In other words, whereas Zhang et al. \cite{zhang2015character} use one-hot vector representations of words or word embeddings, we use binarized pixel values of grayscale images of text corpora.

\begin{table}[!ht]
\centering
\caption{Results of Latin and non-Latin alphabet-based text classification in terms of held-out accuracy. \textit{Worst-Best Performance} reports the results of the worst and best performing baselines from Table 4 of Zhang et al. \cite{zhang2015character} and Conneau et al. \cite{conneau2017very}. Results reported for \textit{TI-CNN} were obtained in 10 epochs}
\resizebox{\linewidth}{!}{\begin{tabular}{|c|c|c|c|c|c|}
\hline
\textbf{Dataset} & \textbf{\begin{tabular}[c]{@{}c@{}}Worst-best\\ Performance\\ (\%)\end{tabular}} & \textbf{\begin{tabular}[c]{@{}c@{}}TI-CNN\\ (\%)\end{tabular}} & \textbf{\begin{tabular}[c]{@{}c@{}}Xception\\ (\%)\end{tabular}}  & \textbf{\begin{tabular}[c]{@{}c@{}}Number of\\ Classes\end{tabular}} \\ \hline
AG's News & 83.1-92.3 & 80.0 & 91.8 &  4 \\
Sogou News (Pinyin) & 89.2-97.2 & 90.2 & 94.6 &  5 \\
Sogou News (Chinese) & 93.1-94.5 & - & \textbf{98.0}  & 5 \\
DBPedia & 91.4-98.7 & 91.7 & 94.5 &  14 \\
Yelp Review Polarity & 87.3-95.7 & 90.3 & 92.8 &  2 \\
Yelp Review Full & 52.6-64.8 & 55.1 & 55.7 & 5 \\
Yahoo! Answers & 61.6-73.4 & 57.6 & 73.0 &  10 \\
Amazon Review Full \footnotemark \label{footnote 1} & 44.1-63 & 50.2 & 57.9 &  5 \\
Amazon Review Polarity \textsuperscript{\ref{footnote 1}} & 81.6-95.7 & 88.6 & 94.0 &  2 \\ \hline
\end{tabular}}
\label{table:text-latin}
\end{table}
\footnotetext{Amazon datasets were large and we did not have enough computational resources to achieve comparable results to SoA with the Xception}

\begin{table}[!ht]
\centering
\caption{\textit{TI-CNN} sentiment prediction for human-generated input text. The model was trained on Amazon Review Polarity dataset}
\begin{tabular}{|c|c|c|}
\hline
\textbf{Sample No.} & \textbf{Text Sample}          & \textbf{Positivity Score}                                                                                                                                            \\ \hline
1                   & 'this product is mediocre'     &      0.60                                                                                                                                      \\
2                   & 'this product is excelent'     &        0.91                                                                                                                                    \\
3                   & 'this product is excellent'      &       0.96                                                                                                                                   \\
4                   & this product is excellent!!!!    &         0.98                                                                                                                                 \\
5                   & 'I love this product it is great'     &    0.99                                                                                                                                 \\
6                   & 'I like this product it is ok'    &       0.78                                                                                                                                  \\
7                   & 'I don't know'                  &           0.56                                                                                                                                \\
8                   & 'as;kdna;sdn nokorgmnsd kasdn;laknsdnaf'   &    0.51                                                                                                                            \\
9                   & \begin{tabular}[c]{@{}c@{}}'I recommended this product to every one in the beginning, \\ but it turned out to be horrible later'\end{tabular}  & 0.64\\
10                  & \begin{tabular}[c]{@{}c@{}}'I recommended this product to every one, \\ in the beginning it was working great, I was in love it'\end{tabular}   &   0.96                              \\ \hline
\end{tabular}
\label{table:human-generated}
\end{table}
%-------------------------------------------------------------------------%

Table \ref{table:text-latin} shows our method's held-out accuracy in the task of Latin and non-Latin (Sogou News in Chinese) alphabet-based text classification for each of the datasets. All baselines are derived from Table 4 of Zhang et al. \cite{zhang2015character} and Conneau et al. \cite{conneau2017very}. We denote the vanilla CNN used by \textit{TI-CNN} (which stands for Text-to-Image Convolutional Neural Networks). The column \textit{Worst-Best Performance} shows the worst and best held-out accuracy achieved by the baseline models. Since this is a 
proof-of-concept of a visual approach to natural language understanding, to further contextualize it, we have also included the equivalent of random classification for each dataset. Our approach achieved comparable results to most of the best performing baselines. For the Yelp Review Full, the results were below the state-of-the-art, since both the \textit{TI-CNN} and Xception were not fine-tuned for each dataset.

Table \ref{table:human-generated} shows human generated text (not included in the training set) the model used for testing. For these examples, the table shows model predictions after the model was trained on the \textit{Amazon Review Polarity} data set \cite{mcauley2013hidden} containing reviews of products in various product categories. The dataset is used for binary (positive/negative) sentiment classification of text and the metric \textit{(positivity score)} is the class probability of being the positive class. The text samples were meant to illustrate our model's merits, compared to pure NLP methods. In detail, the model is able to discriminate between words expressing different degrees of the same sentiment (e.g. samples 1,6 compared to samples 2-5). Sample 2 (compared to samples 3-4) illustrates our method's robustness to anomalies like spelling mistakes. In a traditional NLP setting the misspelled word would have a different representation from the respective correctly-spelled word. Unless the model was trained on data that contained many (and many variations) of these anomalies, or engineered by a human, it would not necessarily correlate the misspelled word with the sentiment it expressed. In our model the misspellings are handled naturally by the network, since the similarity level of misspelled words is high. We note that while this can be alleviated by preprocessing procedures or character-level models, these require more pre-processing or human intervention than our method, which focuses on visual patterns of words. 

As discussed before, the model builds these visual representations in a bottom-up fashion, 
creating a semantic hierarchy which is derived from language use within the context of the corpus domain. 
Sample 4 shows another interesting characteristic of our model which is the understanding of punctuation even if used informally. The exclamation marks used in sample 4 generated the highest prediction score for positive sentiment among all variations of the same phrase (samples 2-4), indicating that the model is capable of understanding the emphasis. Samples 5 and 6 have a similar structure but the different choice of words to describe positive
\begin{figure}[!ht]
    \centering
    \includegraphics[width=0.6\linewidth]{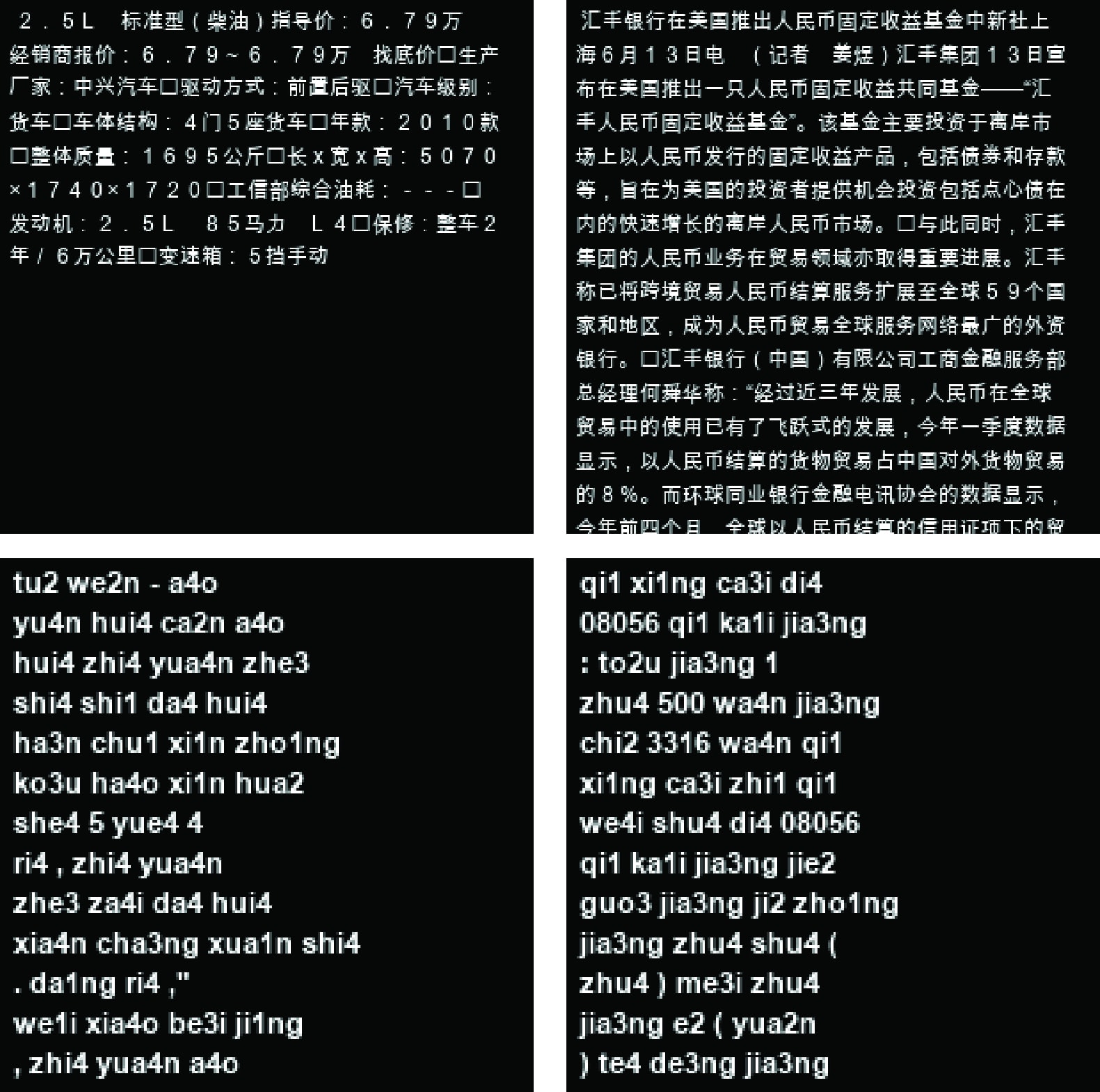}
    \caption{Top: Sogou News dataset with Chinese characters. Bottom: Sogou News dataset with pinyin}
    \label{fig:data_images}
\end{figure}
sentiment affects prediction score. This also exhibits the model's capacity to build meaningful hierarchical representations, as it has learned to discriminate between the small nuances (e.g. choice of words) encountered in (visually and semantically) similar textual structures (sentences). Interestingly, an input which expresses a "neutral" sentiment, such as sample 7, has an analogous prediction score (0.56) that is closer to random guessing in a model that was trained in binary sentiment prediction, which is reasonable behavior. The model is also robust to nonsensical text such as sample 8. The prediction score is, again, close to random guessing.

Samples 9 and 10 are meant to illustrate the model's ability to deal with relatively long sequences of words, which express different (even contradictory) sentiments. Whereas in sample 10 positive sentiment is constantly expressed throughout the sequence, in sample 9 the sentiment switches from positive to negative mid-sequence. While the model erroneously predicts positive sentiment (the prediction score is greater than 0.5), the prediction score is significantly lower for sample 9, compared to sample 10, which indicates that the model has factored in the sentiment switch in its understanding of the text's semantics.

Finally, we applied the Xception architecture to the Sogou News dataset, using the original Chinese
characters (Figure \ref{fig:data_images}). We did not run experiments with the \textit{TI-CNN} for the Sogou News (Chinese), since we only picked the best performing network. Huang and Wang \cite{huang2016character} used 1D CNNs for text classification with Chinese characters and showed that the accuracy recognition was higher than the traditional conversion to the pinyin romanization system. We extended this work by using the Xception architecture in the 2D image to achieve almost the same result (Table \ref{table:text-latin}). This proves that regardless of how many Chinese words we fit in a 300x300 or a 200x200 image, our approach outperformed the NLP sequential CNNs. Furthermore, the performance improved when using the Chinese characters instead of the pinyin.

\subsection{Dialog modeling}
\begin{figure}[!ht]
    \centering
    \includegraphics[width=1\linewidth]{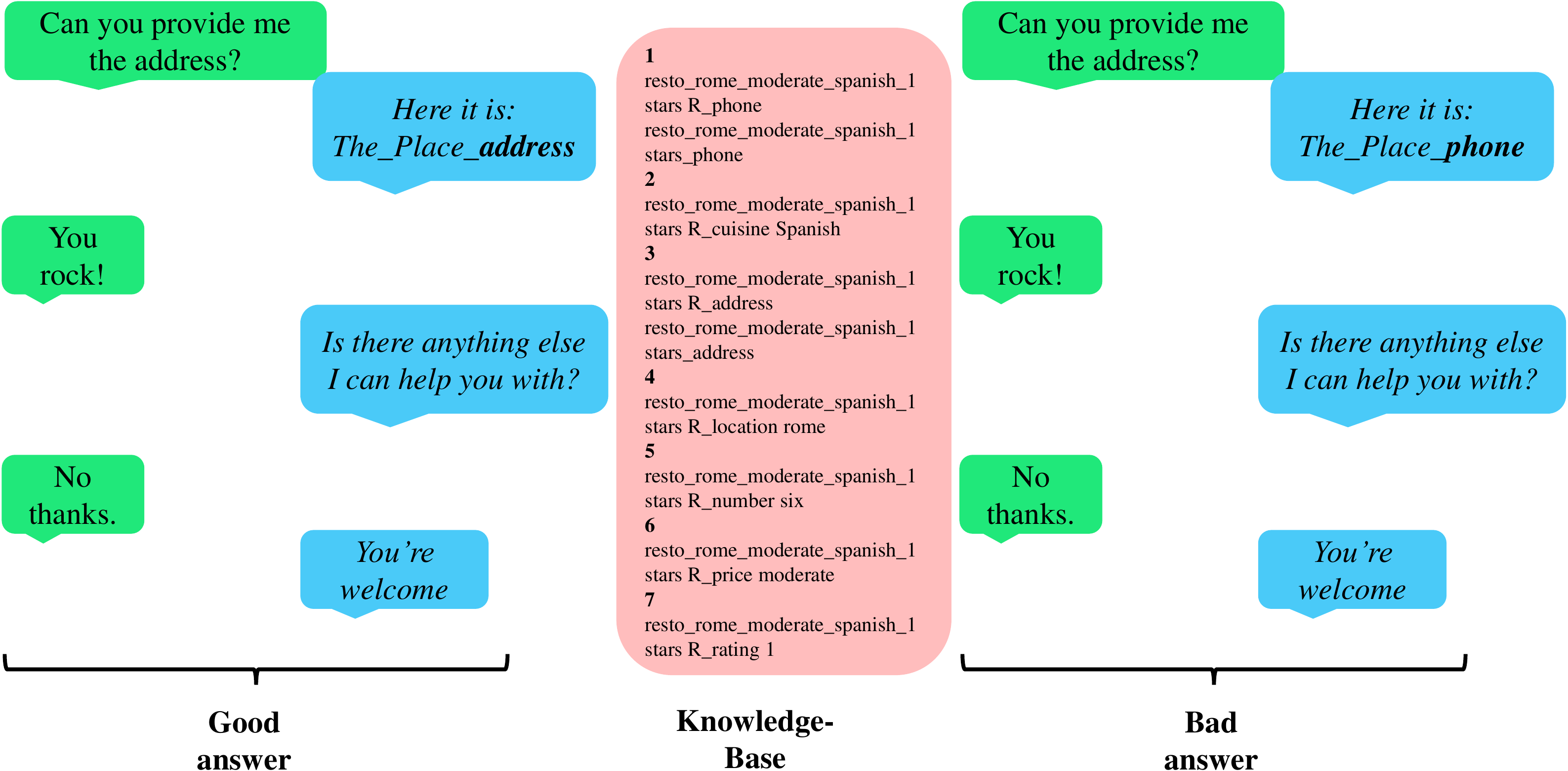}
    \caption{Images of example input data}
    \label{fig:facebook}
\end{figure}

For the dialog modeling task, we tested our Inception-V4-based agent in task 4 (Figure \ref{fig:facebook}) of the bAbI dialog dataset from Bordes et al. \cite{bordes2016learning}. The bAbI dialog dataset consists of 1000 training, 1000 validation and 1000 test 
dialogs in the restaurant booking domain. Each dialog is divided in four different tasks. Here we focus on task 4, 
where the dialog agent should be able to read entries about restaurants from a relevant knowledge base and provide 
the user the information they requested, such as restaurant address or telephone. We note that restaurant telephone 
%-------------------------------------------------------------------------%
%-------------------------------------------------------------------------%
numbers and addresses have been delexicalized and replaced by tokens representing this information. We chose to 
focus on this task to demonstrate the increased effectiveness of visual processing of dialog as opposed to purely 
linguistic processing, due to the high number of different lexical tokens. In our approach the agent needs to correlate 
the visual pattern of a knowledge base entry to the relevant request. While in principle this should be easy to 
achieve using artificial delexicalized tokens, as in this benchmark task, it would be far more difficult to do so in 
the real world, with non-standard sequences of words (such as restaurant names, addresses, telephone numbers, etc). 
However, given the results of the text classification tasks, we hypothesize that given enough data, our visual approach 
can create semantic models that encapsulate such correlations.

As explained in 
Section \ref{sec:models}, we used Inception-V4, since our original architecture was too simple for the task of dialog 
modeling and was under-performing. As such, we used a modern powerful architecture to overcome these issues. 
Additionally, we introduced data augmentation methods inspired by both modalities, which included random character flipping and chose textually similar candidate answers, rotations and shifts of the width and height of the input image. This resulted in a variety of (visually) wrong candidate 
responses which we added to the candidate list. Our augmentation methods, which are different from traditional image data augmentation methods, improved the performance of the Inception-V4. 

As in text classification, we trained 
the model with images of dialog text taken from the bAbI corpus. So the agent learns the expected user utterances 
and their corresponding responses on the system side by processing images of in-domain dialog text. The agent learned 
visual representations of text meaning and structure both at word-level (implicitly, through the optimization process) 
and utterance-level (explicitly, through labeling of correct and incorrect responses given a user utterance). 

\begin{table}[!ht]
\centering
\caption{Facebook bAbI Dialog Task 4}
\resizebox{\linewidth}{!}{
\begin{tabular}{|c|c|c|c|c|}
\hline
\textbf{Metrics} & \textbf{\begin{tabular}[c]{@{}c@{}}Inception-V4\\ (\%)\end{tabular}} & \textbf{\begin{tabular}[c]{@{}c@{}}Inception-V4\\ w/ Augmentation\\ (\%)\end{tabular}} & \textbf{\begin{tabular}[c]{@{}c@{}}Memory Networks\\ w/o Match Type\\ (\%)\end{tabular}} & \textbf{\begin{tabular}[c]{@{}c@{}}Memory Networks\\ w/ Match Type\\ (\%)\end{tabular}} \\ \hline
Per-response Accuracy & 63.3 & 99.8 & 59.5 & 100.0 \\
Per-dialog Accuracy & 11.4 & 99.6 & 3.0 & 100.0 \\

\hline
\end{tabular}}
\label{table:fb-dial}
\end{table}

Table \ref{table:fb-dial} shows the Inception-V4 performance against the memory networks (MemNets) used in \cite{bordes2016learning}. 
Our augmentation method improved the response accuracy by approximately 57.66\%, and increased dialog 
completion rate from 11.4\% to 99.6\% for dialogs that contained words and tokens encountered in the training data. This high recognition accuracy increase is due to the fact that even without augmentation, the correct answer was within the top 10 
textually similar wrong candidate answers (one character was different between correct and wrong answer). With augmentation, harder dialogues are introduced 
into the training set, enabling the network to differentiate smaller details.

The table also shows that our approach is 
competitive with MemNets, even without augmentation, when the latter does not use match types. Bordes et al. \cite{bordes2016learning} introduced match types to make their model rely on type information, rather than exact match of word embeddings 
corresponding to words that frequently appear as containing out of vocabulary (OOV) entities (restaurant name, telephone number, etc). This is because it is 
hard to distinguish between similar embeddings in a low-dimensional embedding space (\textit{e.g.} telephone 
numbers) as they lead to full scores in all metrics. However, as discussed above, 
we believe that if pre-trained on images of similar in-domain non-dialog text corpora, CNN models can deal with 
such phenomena in a natural way.

%-------------------------------------------------------------------------%
\section{Conclusion}
We presented a proof-of-concept for natural language understanding, relying only on visual 
features of text. For non-dialog text, images of text as input to the CNN models can build hierarchical semantic representations which let them detect various subtleties in language use, irrespective of the language of the input data. For dialog text, we showed that CNN models learn both the structure of discourse and the implied dialog pragmatics implicitly contained within the training data. Crucially, our approach does not require any preprocessing of natural language data, traditionally found in NLP applications, such as tokenization, optical character recognition, stemming, or spell checking. Our method can work using different computer fonts, background colors and can be expanded to human handwriting. Unlike traditional and language agnostic models, our method can perform NLP tasks on real-word documents that include tables, bold, underlined and colored text, where traditional NLP methods, as well as language agnostic models (1D CNN) fail.

Our work is a first step towards expanding the methods for natural 
language understanding, exploiting recent advances in image recognition and computer vision. Initial results 
of this approach are promising for a wide range of NLP tasks, such as text classification, sentiment analysis, 
dialog modeling and natural language understanding. Future work will study the effect of pre-trained 
models on non-dialog corpora with regard to modeling performance, incorporation within generative frameworks for 
text generation for tasks like text summarization or performance of more complex applications (such as recurrent CNNs). For the task of text classification, the recognition accuracy of the Xception will increase, since the reported results are achieved without any fine-tuning for the specific datasets. As computer vision deep learning models continue to improve, we expect our results for the NLP task to follow suit (Table \ref{table:text-latin}).

\medskip

\small

\bibliographystyle{plain}
\bibliography{egbib}

\end{document}